\documentclass[a4]{article} 

\usepackage{PRIMEarxiv}

\usepackage[utf8]{inputenc} 
\usepackage[T1]{fontenc}    
\usepackage{hyperref}       
\usepackage{url}            
\usepackage{booktabs}       
\usepackage{amsfonts}       
\usepackage{nicefrac}       
\usepackage{lipsum}
\usepackage{fancyhdr}       
\usepackage{graphicx}       
\graphicspath{{media/}}     

\usepackage{algorithm}
\usepackage{algpseudocode}
\usepackage{natbib}
  \setcitestyle{aysep={}}

\renewcommand{\cite}[1]{\citep{#1}}
\newcommand{\bm}[1]{{\mbox{\boldmath $#1$}}}
\newcounter{bean} \newcounter{subbean}

\title{Automatic Construction of Pattern Classifiers Capable of Continuous Incremental Learning and Unlearning Tasks Based on Compact-Sized Probabilistic Neural Network%
\thanks{A modified version appeared in the Proceedings of the AAIML2026.}
}
\author{
  Tetsuya Hoya and Shunpei Morita\\
  Department of Computer Engineering, College of Science \& Technology\\
  Nihon University\\
  Funabashi, Chiba, JAPAN\\
  \texttt{houya.tetsuya@nihon-u.ac.jp}\\
}

\begin{document}
\maketitle
\begin{abstract}
This paper proposes a novel approach to pattern classification using a probabilistic neural network model.  The strategy is based on a compact-sized probabilistic neural network capable of continuous incremental learning and unlearning tasks.  The network is constructed/reconstructed using a simple, one-pass network-growing algorithm with no hyperparameter tuning.  Then, given the training dataset, its structure and parameters are automatically determined and can be dynamically varied in continual incremental and decremental learning situations.  The algorithm proposed in this work involves no iterative or arduous matrix-based parameter approximations but a simple data-driven updating scheme.  Simulation results using nine publicly available databases demonstrate the effectiveness of this approach, showing that compact-sized probabilistic neural networks constructed have a much smaller number of hidden units compared to the original probabilistic neural network model and yet can achieve a similar classification performance to that of multilayer perceptron neural networks in standard classification tasks, while also exhibiting sufficient capability in continuous class incremental learning and unlearning tasks.
\end{abstract}

\section{Introduction}
In general, one of the central issues in machine learning is the selection of algorithmic and model hyperparameters.  The selection is nontrivial in the case of deep learning (DL) \cite{LeCun-2015} approaches since the number of hyperparameters is typically large, i.e., the number of hidden layers and units per hidden layer, the batch size, the number of epochs, the learning constant, the choice of the loss function, and other algorithm-dependent parameters.  These parameters can also significantly affect the performance and generally have to be determined through multiple simulation runs on a trial-and-error basis.  Moreover, each run involves iterative parameter tuning of the model and often suffers from inherent ill-posed problems such as gradient vanishing/exploding and local minima.  Hence, training of DL-oriented models is usually quite resource-demanding and time-consuming.

On the other hand, it is also known that standard deep neural network (DNN) \cite{Schmidhuber-2015}, a.k.a. multilayer perceptron neural network (MLP-NN) with multiple hidden layers \cite{Rumelhart-1986}, approaches incur the so-called `catastrophic forgetting' \cite{McCloskey-1989} in the case of incremental learning; newly given training data will corrupt the data space represented by a DNN.  One of the representative and practical approaches to avoid the data space corruption problem is based on the so-called `replay' (or rehearsal) method \cite{Hetherington-1989}, such as the iCaRL \cite{Sylvestre-2017}, where the network is retrained using a subset of the previously seen training data and new ones.  However, since replay-based methods still require storing (at least some portion of) the previously used training data, it is said that this approach does not yield the proper solution to the incremental learning tasks.  

For unlearning, several methods for DNNs have been proposed to date \citep[cf.][]{Bourtoule-2021,Kodge-2023,Tarun-2023}, and they attempt to remove the learned information from the static network.  Hence, this approach contrasts the proposed scheme, where the network structure dynamically varies according to the situation.  Moreover, they generally require additional complexity to the model approximation (e.g., another weight updating that involves a series of matrix operations and/or storing several snapshots of the model) and, therefore, are computationally resource-demanding.  To the author's knowledge, no unified DNN-based scheme capable of effectively performing continual incremental learning and unlearning tasks has been proposed.

In contrast to DNNs, it has been shown that incremental learning tasks can be straightforwardly performed using a probabilistic neural network (PNN) \cite{Specht-1990,Hoya-2003,Takahashi-2022,Morita-2023}.  A PNN is a three-layer network with radial basis function (RBF) units in the hidden layer and linear output layer units, unlike DNNs.  The PNN's capability of straightforward incremental learning owes to its inherent characteristics of transparent architecture and local data representation, while the data learned are distributed throughout the network parameters in the DNNs.

This paper proposes a simple pattern classification scheme using PNNs requiring no hyperparameter tuning.  Given the training dataset, the network size (i.e., corresponding to the number of hidden and output units) is automatically determined by applying the proposed construction algorithm, and it can be significantly smaller than that of the original PNN; in the case of the original PNN, all the training data must be accommodated within the hidden layer, and this manner of accommodation may often cause the problem of over-training.  The present work also reveals that the compact-sized PNN (CS-PNN) can sufficiently deal with continual incremental learning and unlearning situations by dynamically varying the network size while keeping reasonable classification performance.
\subsection{Incremental Learning and Unlearning Tasks}\label{sec:tasks}
The scheme proposed in this paper can be applied to both instance- and class-wise tasks for incremental learning and unlearning situations.  These two tasks differ in the available training data for restructuring the pattern classifier during continual learning/unlearning.  In the instance-wise incremental learning tasks (IIL), the pattern classifier is updated using each batch containing the training data for all or partial classes.  In contrast, only the data for a group of classes are available at a time for class-wise (or class) incremental learning (CIL).  Therefore, the CIL is generally considered more challenging than IIL, as the pattern classifier needs to reestimate the pattern space for the unknown classes.  On the other hand, instance-/class-wise unlearning (a.k.a. class decremental learning (CDL), for the latter) considers situations where the undesired or malfunctioning samples of the same/multi-classes learned by the classifier must be removed.
\section{Method}\label{sec:method}
\subsection{Probabilistic Neural Network}\label{sec:PNN}
PNN \cite{Specht-1990} is a variant of radial basis function neural networks (RBF-NNs) \cite{Broomhead-1988}.  It requires much less training effort than typical DL approaches: training of a PNN is completed by simply assigning each training data to the centroid vector of an RBF $\bm{c}_j$ $(j=1, 2, \ldots, N_h)$ in the hidden layer, as given by (\ref{eqn:rbf}).

\begin{equation}
  h_j(\bm{x}) = \exp(-\frac{\|\bm{x}-\bm{c}_j\|^2_2}{\sigma^2})
  \label{eqn:rbf}
\end{equation}

Then, the unique hyperparameter of the radius value for each RBF is manually specified to complete the training phase.  However, the PNN so obtained tends to suffer from slow testing (reference) mode and over-fitting, as all the training data need to be accommodated within the network.  It even becomes prohibitive in practice where the training dataset is immense.  To circumvent these problems inherent to PNN, a data-clustering approach, such as $k$-means  \cite{MacQueen-1967}, to the training dataset is generally performed before the PNN training.  However, the application of data-clustering methods typically adds another complexity to the hyperparameter choice, i.e., in the case of PNN, the number of the training data to be accommodated within the network (i.e., corresponding to the number of the hidden units) and other algorithmic parameters (i.e., the seed value for the random choice of the initial clusters and the threshold value to judge reaching a state of convergence for updating the centers, in the case of the $k$-means; cf. \citet{Takahashi-2022}).

\begin{figure}[h]
  \centering
  \includegraphics[width=\textwidth]{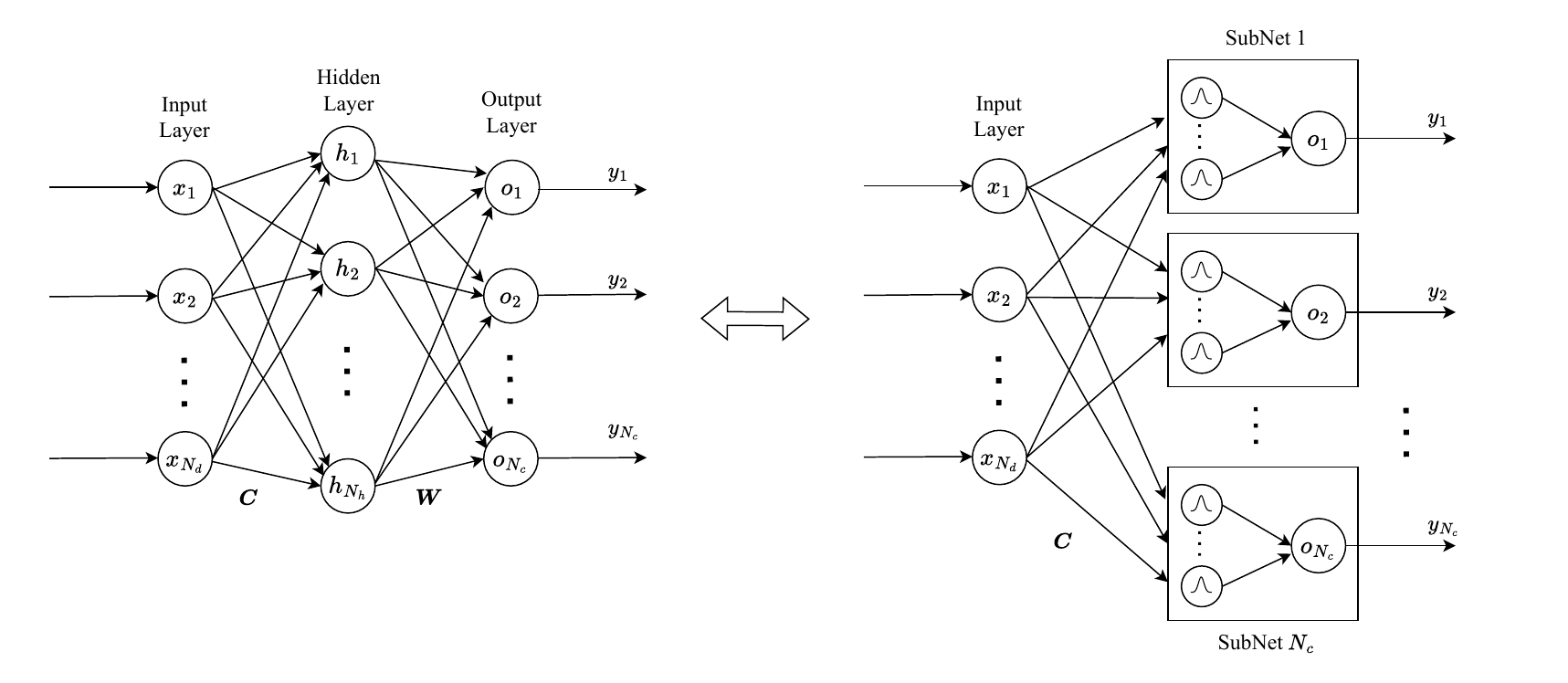}
  \caption{A PNN (left) and its topologically equivalent structure with $N_c$ subnets (right) \cite{Hoya-1998}.  In the figure, the matrices $\bm{C}=[\bm{c}_1, \bm{c}_2, \ldots, \bm{c}_{N_h}]$ and $\bm{W}=\{w_{jk}\}$ where $w_{jk}=0/1$; all the input units $x_i$ $(i=1, 2, \ldots, N_d)$ are connected to each hidden layer unit $h_j$ $(j=1, 2, \ldots, N_h)$ with the respective weight values denoted by the elements in $\bm{c}_j$.  Each of $h_j$ is connected only to one of the output units $o_k$ $(k=1, 2, \ldots, N_c)$ with the weight values $w_{jk}=1$ (if it belongs to the same class as $o_k$) and $0$ (otherwise).}
  \label{fig:fig1}
\end{figure}

Unlike ordinary DNN models, a standard PNN has a shallow, three-layered architecture; between the input and output layer, there is only a single hidden layer comprised of multiple RBF nodes.  Each RBF node is connected to all the input nodes with weight values equal to the respective element values of the centroid vector.  In contrast, an RBF node of a PNN is connected only to a single output node corresponding to the same class with the connection weight unity.  Therefore, the hidden-to-output layer part of a PNN is topologically equivalent to a collection of subnets \cite{Hoya-1998}, each responsible for a single class, as shown in the right part in Fig. \ref{fig:fig1}.
Since each subnet is distinct, class-wise network growing and shrinking is straightforward from the structural viewpoint, in contrast to the DNN models.  However, to perform network growing/shrinking, the unique value of the radius for each RBF needs to be adjusted according to the pattern space so modified, and this adjustment is generally nontrivial.
\subsection{The Unique Radius}\label{sec:radius}
Typically, though the value of the unique radius of a PNN is heuristically chosen, the radius setting in the following is suggested in the authors' previous work \cite{Morita-2023}:

\begin{equation}
  \sigma = \frac{D_{max}}{N_c}
  \label{eqn:radius1}
\end{equation}

where $N_c$ is the number of classes, and $D_{max}$ is the maximal distance found among all the pairs of training patterns.  In (\ref{eqn:radius1}), the entire $d$-dimensional pattern space ($\mathbb{R}^d$) is assumed to comprise the distinct hypercubes of an equal side $r$ for each class and be covered by a hypersphere with the radius $r=D_{max}/N_c$.  However, the setting by (\ref{eqn:radius1}) is considered effective only in ordinary, \emph{static} pattern classification situations, where $N_c$ is known \emph{a priori} and all the training data are available.  On the other hand, there can be other situations in practice where the data are only available for partial classes at an earlier stage of a pattern classification task and where the number of classes is dynamically varied due to some environmental changes.  Therefore, in such an online situation, the system has to deal with the continuously varying pattern space.
\subsection{The Compact-Sized PNN (CS-PNN)}\label{sec:CSPNN}
Also, the primary focus of the previous work \cite{Morita-2023} was to construct a pattern classifier capable of performing class-incremental learning (CIL) tasks based on a PNN with fewer RBFs than the original model, a.k.a.  a CS-PNN.  However, the approach has inherent drawbacks: i) the maximal distance $D_{max}$ in (\ref{eqn:radius1}) has to be computed in advance using all the training pairs and \emph{fixed} while performing the CIL.  Hence, this setup is not considered fully practical;  ii) there is a hyperparameter $\theta$, i.e., a threshold value to judge whether a new RBF is added for the construction algorithm, and the hyperparameter selection still depends upon some heuristics.
\subsection{Proposed Method}\label{sec:proposed}
This work, therefore, proposes a novel method for a CS-PNN to cope with the drawbacks of the original approach described above.  First, unlike the ordinary pattern classification scheme, a varying value of the unique radius is used for each RBF in both the training (construction) and testing (reference) modes within the proposed method: the value of $d_{max}$ within (\ref{eqn:radius2}) is updated to track the varying pattern space whenever new data arrives to the classifier.  By so doing, the pattern classifier is expected to handle the incremental learning (i.e., reconstruction) and the unlearning situations effectively.  Second, within the construction/reconstruction algorithm, a new RBF is added if incoming training data is incorrectly classified, rather than when the activation of any already accommodated RBFs is below a manually given threshold as in the previous approach \cite{Morita-2023}; \emph{this modification eventually leads to removing the necessity of the hyperparameter selection}.  Therefore, once a training dataset is given, the pattern classifier is automatically constructed without any hyperparameter choice. 

The pseudocodes of the proposed method for construction/reconstruction (both for the IIL and CIL tasks), unlearning (for the instance-wise unlearning and CDL tasks), and testing modes are shown in \textbf{Algorithms \ref{alg-construction} - \ref{alg-testing}}, respectively.

As equation (\ref{eqn:radius2}) in \textbf{Algorithm \ref{alg-construction}} indicates, the radius varies with the current number of classes $k$ accommodated within the CS-PNN.  Simultaneously, the numerator $d_{max}$ is updated to keep track of the modified pattern space.

It is noticeable that, within both the construction/reconstruction and testing algorithms (shown as \textbf{Algorithms \ref{alg-construction}} and \textbf{\ref{alg-testing}}, respectively), the computation of $d_{max}$ is conveniently carried out \emph{in parallel to} the feed-forwarding of the input data to each existing RBF of a PNN; the arithmetic operation of finding a maximal value proceeds, while the distance between the input and each centroid vector (i.e., represented by the squared $L2$-norm in the numerator within the exponential function in (\ref{eqn:rbf})) is computed.  Therefore, the additional computational complexity for $d_{max}$ is negligible.  Note also that, for the unlearning situations (\textbf{Algorithms \ref{alg-unlearning} and \ref{alg-cdl}}), the RBF units or subnets responsible for no longer-used classes are just unloaded from the PNN without involving a further computation of $d_{max}$; the value will be automatically updated during the testing mode afterward.
\begin{algorithm}[tb]
\caption{Construction/Reconstruction of a CS-PNN (for both CIL and IIL tasks)}
\label{alg-construction}
\begin{algorithmic}[1]
\State \textbf{input:} $\bm{X}_{tr}=\{(\bm{x}_{tr}(1), t_{tr}(1)), (\bm{x}_{tr}(2), t_{tr}(2)), \ldots, (\bm{x}_{tr}(N_{tr}), t_{tr}(N_{tr}))\}$ ($\bm{x}_{tr}(i)$: training pattern vector, $t_{tr}(i)$: target class label)
\If {[Initial Construction]}
  \State $k \gets 1$, add $o_1$, set $\phi(o_1) \gets t_{tr}(1)$ ($\phi(o_k)$: class label for $o_k$)
  \State $j \gets 1$, add $h_1$ with $\bm{c}_1 \gets \bm{x}_{tr}(1)$ and $w_{11} \gets 1$
  \State $s \gets 2$
\Else
  \State $s \gets 1$
\EndIf
\For {$i=s$ to $N_{tr}$}
  \If {there is no output unit corresponding to $t_{tr}(i)$}
    \State $k \gets k+1$, add $o_k$, set $\phi(o_k) \gets t_{tr}(i)$
    \State $j \gets j+1$, add $h_j$ with $\bm{c}_j \gets \bm{x}_{tr}(i)$ and $w_{jk} \gets 1$
  \Else
    \State Find the maximal distance $d_{max}$ among all the pairs of $\bm{x}_{tr}(i)$ and $\bm{c}_j$
    \State Set the unique radius for each $h_j$:
    \State 
      \begin{equation}
        \sigma \gets \frac{d_{max}}{k}
        \label{eqn:radius2}
      \end{equation}
    \State Compute $h_j(\bm{x}_{tr}(i))$
    \State Compute $o_k$, find the maximal: $K \gets arg max(o_k)$
    \If {$\phi(o_K) \neq t_{tr}(i)$}
      \State $j \gets j+1$, add $h_j$ with $\bm{c}_j \gets \bm{x}_{tr}(i)$ and $w_{jK} \gets 1$
    \Else
      \State Find a maximally activated RBF within SubNet $K$: $J \gets arg max(h_j)$
      \State Update $\bm{c}_J$:
      \begin{equation}
        \bm{c}_J \gets \frac{\bm{c}_J+\bm{x}_{tr}(i)}{2}
      \end{equation}
    \EndIf
  \EndIf 
\EndFor
\State \textbf{output:} updated PNN
\end{algorithmic}
\end{algorithm}
\begin{algorithm}[tb]
\caption{Unlearning of a CS-PNN (for instance-wise unlearning tasks)}
\label{alg-unlearning}
\begin{algorithmic}[1]
\State \textbf{input:} IDs of the hidden (RBF) units to remove: \{$p(1), p(2), \ldots, p(N_j)$\}
\For {$i=1$ to $N_j$}
  \State Remove the RBF units $h_{p(i)}$ and the corresponding input-hidden layer connections $\bm{c}_{p(i)}$ from a CS-PNN
\EndFor
\State $j \gets j - N_j$
\State \textbf{output:} updated PNN
\end{algorithmic}
\end{algorithm}
\begin{algorithm}[tb]
\caption{Unlearning of a CS-PNN (for CDL tasks)}
\label{alg-cdl}
\begin{algorithmic}[1]
\State \textbf{input:} class labels to remove: \{$l_1, l_2, \ldots, l_{N_r}$\}
\For {$i=1$ to $N_r$}
  \State Remove the SubNet corresponding to the class label $\phi(i)=l_i$ from a CS-PNN
\EndFor
\State $k \gets k - N_r$
\State \textbf{output:} updated PNN
\end{algorithmic}
\end{algorithm}
\begin{algorithm}[tb]
\caption{Testing of a CS-PNN}
\label{alg-testing}
\begin{algorithmic}[1]
\State \textbf{input:} $\bm{X}_{ts}=\{(\bm{x}_{ts}(1), t_{ts}(1)), (\bm{x}_{ts}(2), t_{ts}(2)), \ldots, (\bm{x}_{ts}(N_{ts}), t_{ts}(N_{ts}))\}$ ($\bm{x}_{ts}(i)$: testing pattern vector, $t_{ts}(i)$: target class label)
\For {$i=1$ to $N_{ts}$}
  \State Find the maximal distance $d_{max}$ among all the pairs of $\bm{x}_{ts}(i)$ and $\bm{c}_j$
  \State Set the unique radius $\sigma$ for each $h_j$ according to (\ref{eqn:radius2}), compute $h_j(\bm{x}_{ts}(i))$
  \State Compute $o_k$, find the maximal: $K \gets arg max(o_k)$
  \If {$\phi(o_K)=t_{ts}(i)$}
    \State [Correctly Classified]
  \EndIf
\EndFor
\State \textbf{output:} classification results of $\bm{X}_{ts}$
\end{algorithmic}
\end{algorithm}
\section{Simulation Study}\label{sec:simulation}
Nine publicly available datasets, eight from the UCI machine learning repository \cite{Dua-UCI} and the MNIST database \cite{LeCun-MNIST}, were used to conduct the simulation study.  The datasets selected for the simulation study are used for a variety of classification tasks, including isolated alphabetic letter speech (isolet) / handwritten digit (optdigits and pendigits) / alphabet (letter-recognition) image or radar data (ionosphere) recognition. Table \ref{table:datasets} summarizes the datasets used for the simulation study.  As in Table \ref{table:datasets}, the condition varies per dataset, i.e., the number of samples, classes, or features.  Each sample in the datasets was normalized within the range [-1.0 1.0], and no other preprocessing was made before performing the simulations.

For the simulation study, three tasks were considered: standard classification, class incremental learning, and continuous multi-class unlearning and incremental learning, as described next.
\begin{table}
 \caption{Summary of the datasets used for the simulation 
    study}
  \centering
 \begin{tabular}{lllll}
    \toprule
    Dataset & \#Training & \#Testing & \# Classes 
      & \#Features per pattern \\
    \midrule
    abalone    & 3133 & 1044 & 3 & 10\\
    ionosphere & 200  & 151  & 2 & 34\\
    isolet     & 6238 & 1559 & 26 & 617\\
    letter-recognition & 16000 & 4000 & 26 & 16\\
    MNIST      & 60000 & 10000 & 10 & 784\\
    optdigits  & 3823 & 1797 & 10 & 64\\
    pendigits  & 7494 & 3498 & 10 & 16\\
    sat        & 4435 & 2000 & 6  & 36\\
    segmentation & 210 & 2100 & 7 & 19\\
    \bottomrule
  \end{tabular}
  \label{table:datasets}
\end{table}
\subsection{Standard Classification Tasks}
In the simulation study, ordinary pattern classification tasks (i.e., standard classification tasks) were first considered: each training sample was available for all the classes in a dataset.  Therefore, performing the standard tasks provided a baseline for each pattern classifier, and three pattern classifiers were considered for the standard tasks: i) original PNN, ii) CS-PNN, and iii) MLP-NN.  For training an MLP-NN, the Adam algorithm \cite{Kingma-2015} was chosen.

Although the CS-PNN does not require any hyperparameter setting in advance, as described above, the original PNN has a single hyperparameter of unique radius to be given before performing the pattern classification tasks; the radius setting given by (\ref{eqn:radius1}) \cite{Morita-2023} was used in the simulations.  In contrast to the PNN approaches, the number of hidden layers and units per hidden layer has to be set in advance for MLP-NNs.  In the present work, the number of hidden layers was fixed at $N_{hL}=2$ for an MLP-NN.  Then, the number of units in each hidden layer was given according to the following setting:

\begin{equation}
  N_h = \frac{2}{3} (N_d+N_c)~,
  \label{eqn:nhunits}
\end{equation}

where $N_d$ and $N_c$ are the number of input units (i.e., equal to the number of features per pattern) and that of classes (i.e., same as the number of the output layer units), respectively.

There are also many algorithm-dependent hyperparameters for the MLP-NN that need to be determined \emph{a priori}.  During the preliminary simulation study, although parameters such as the batch size ($N_{batch}$), number of epochs ($N_{epochs}$), learning constant ($\alpha$), and two betas ($\beta_1$ and $\beta_2$) for the Adam were found to affect the performance significantly, the settings $N_{batch}=128$ and $N_{epochs}=20$ and those for the Adam similar to the default ones used in PyTorch library (\url{https://pytorch.org/}) gave a moderately reasonable performance, i.e., $\alpha=0.001$ or $0.01$, $\beta_1=0.9$, and $\beta_2=0.999$.

Table \ref{table:sim1} summarizes the simulation results of the standard tasks.
\begin{table}
 \caption{Summary of the simulation results for the standard tasks}
  \centering
   \begin{tabular}{lcccccc}
    \toprule
     & \multicolumn{2}{c}{Original PNN} & \multicolumn{2}{c}{CS-PNN} & \multicolumn{2}      {c}{DNN ($N_{hL}=2$)}\\
    \cmidrule(r){2-3} \cmidrule(r){4-5}\cmidrule(r){6-7}
    Dataset & Acc. (\%) & $N_h$(=\#Training) & Acc. (\%) & $N_h$ & Av. Acc. (\%) & $N_h \times N_{hL}$\\
    \midrule
    abalone & 53.35 & 3133 & 52.78 & 1114 & \textbf{54.46} & 18 \\
    ionosphere & 85.43 & 200 & \textbf{90.07} & 92 & 75.76 & 48 \\
    isolet & 88.84 & 6238 & 87.94 & 1327 & \textbf{95.05} & 858 \\
    letter-recognition & \textbf{96.22} & 16000 & 92.45 & 2043 & 72.93 & 56 \\
    MNIST & 96.50 & 60000 & 94.90 & 3684 & \textbf{98.02} & 1058 \\
    optdigits & \textbf{98.39} & 3823 & 95.05 & 188 & 94.85 & 98 \\
    pendigits & 94.25 & 7494 & \textbf{95.05} & 263 & 90.35 & 34 \\
    sat & 81.15 & 4435 & 80.30 & 403 & \textbf{84.51} & 56 \\
    segmentation & \textbf{85.33} & 210 & 82.81 & 49 & 33.07 & 34 \\
    \bottomrule
  \end{tabular}
  \label{table:sim1}
\end{table}
In Table \ref{table:sim1}, note that the classification accuracy averaged over ten runs for each dataset is shown for the DNN (MLP-NN) since the performance varied with random initialization of the network parameters for each run.  Also, note that the number of hidden units in the original PNN used equals that of the training patterns for each dataset, as shown in the second column in Table \ref{table:datasets}.

As shown in Table \ref{table:sim1}, the classification performance obtained using the CS-PNN was more or less comparable to that of the original PNN with fewer hidden units (i.e., from around 4 to 46\% of the units required). In contrast, the number of hidden units generated in the CS-PNN was much larger than that of the DNN, while the performance of the DNN dropped largely for some datasets.  During the simulations, however, it was noticed that the performance of the DNN varied significantly and was even unstable in some cases.  One of the reasons for this is probably due to the small number of hidden units used; however, identifying the specific reason was not possible since many other hyperparameters, as described earlier, were inherently involved in training the DNN, and a particular coherent manner of the selection is yet to be found.
\subsection{Class Incremental Learning Tasks}
For the CIL tasks, the number of classes for the partial data given at a time to perform incremental learning varied from one to four.  Then, a pattern classification task for the newly added and already learned classes was performed for each task.

In the simulation study, seven datasets out of nine with more than five classes, as shown in Table 1, were used: a) isolet, b) letter-recognition, c) MNIST, d) optdigits, e) pendigits, f) sat, and g) segmentation.  The performance obtained using CS-PNN for each dataset was compared with a DNN with two hidden layers using the iCaRL method \cite{Sylvestre-2017}.  The iCaRL was chosen for comparison since the replay method has been used widely for DNNs in CIL situations \cite{Masana-2023}.  For training the DNN, the same settings for the Adam algorithm for the standard classification tasks were used.  In contrast, the memory size for the replay by the iCaRL, i.e., the number of training data stored for the subsequent CIL tasks, was set as 0.2 times the number of all the training data available in each dataset \cite{Morita-2023}.

\begin{figure}[h]
  \centering
  \includegraphics[width=\textwidth]{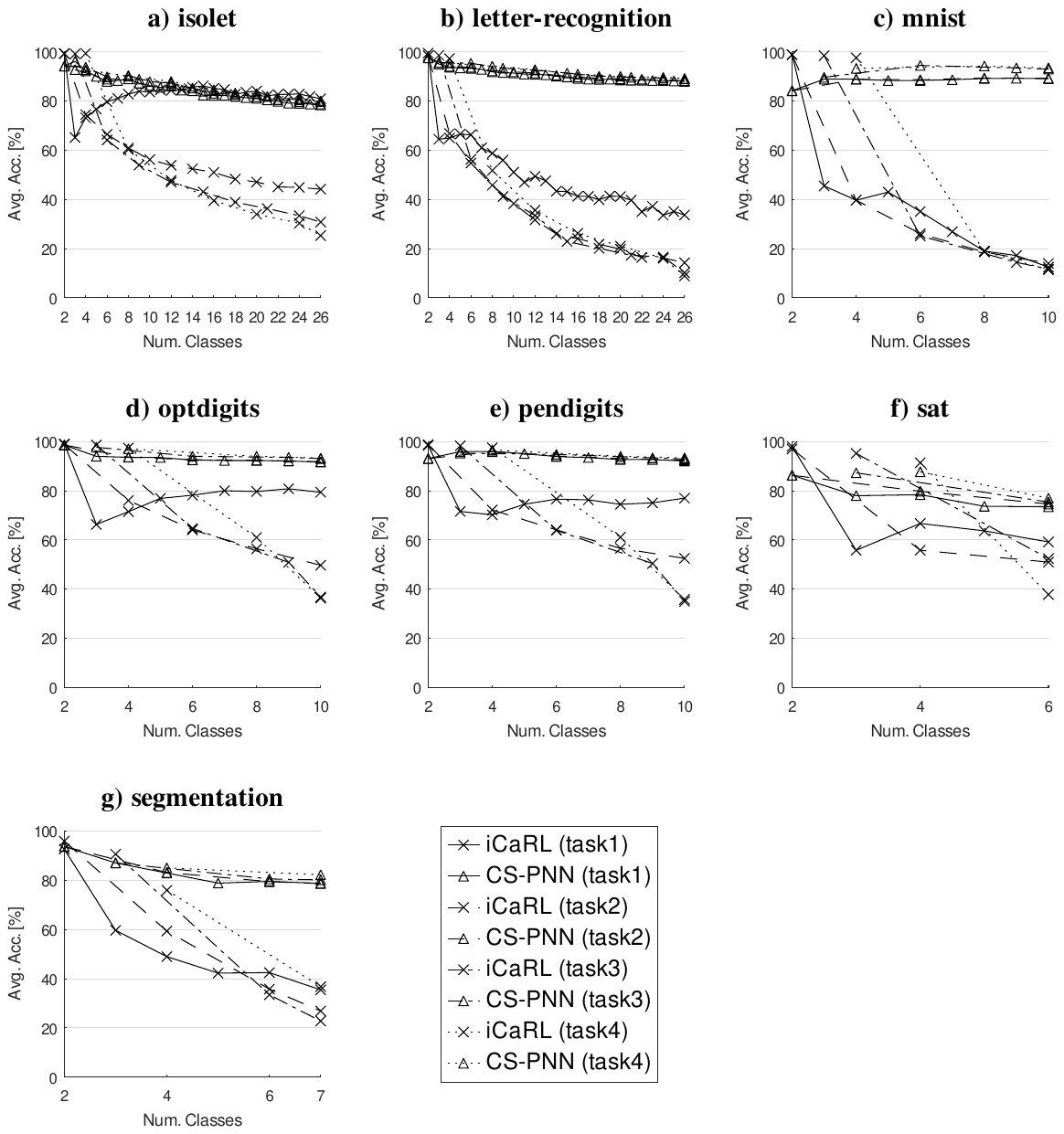}
  \caption{Transition of the classification accuracy (averaged) for the CIL tasks using the seven datasets: a) isolet, b) letter-recognition, c) MNIST, d) optdigits, e) pendigits, f) sat, and g) segmentation.}
  \label{fig:sim21}
\end{figure}

\begin{figure}[h]
  \centering
  \includegraphics[width=\textwidth]{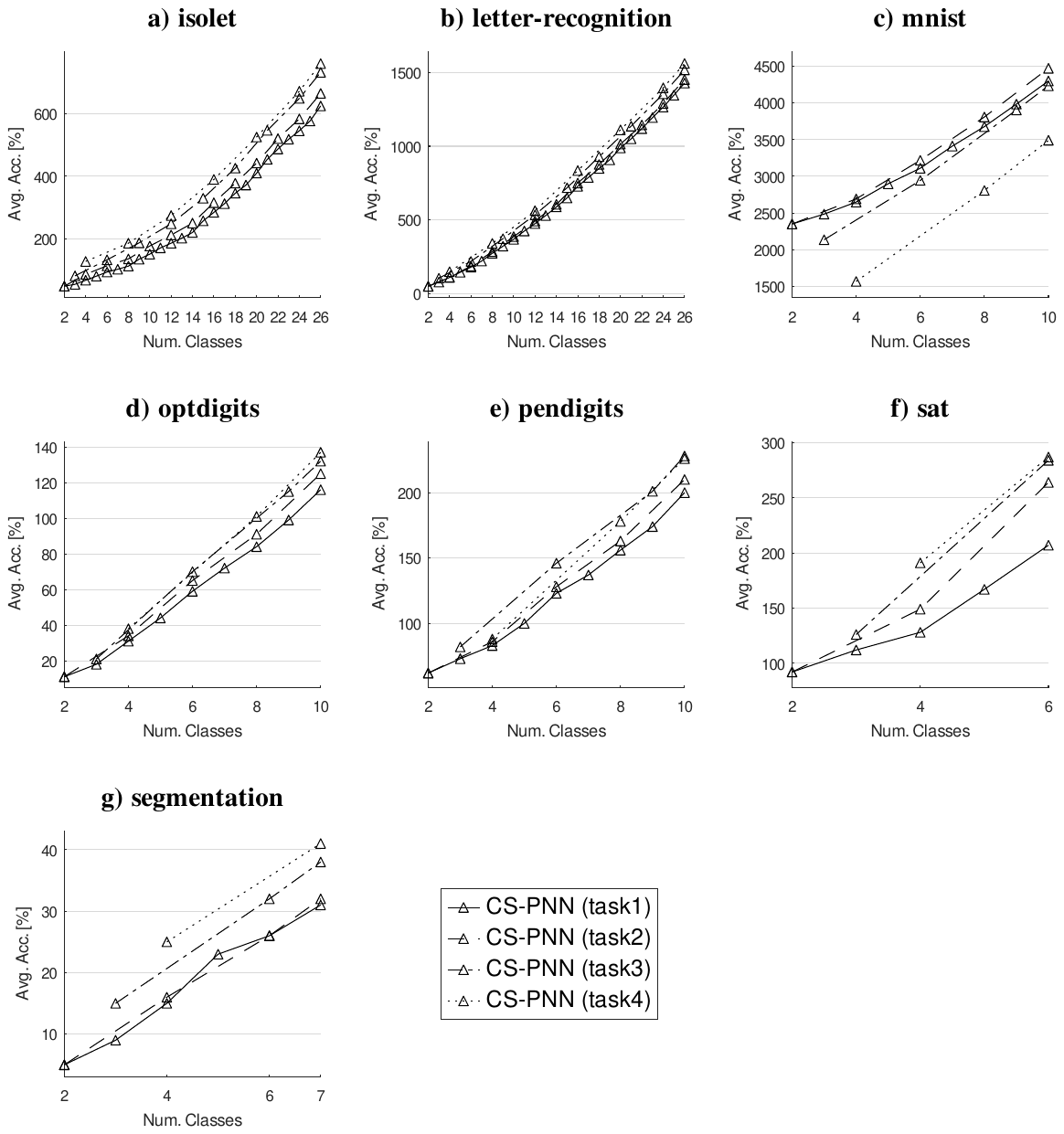}
  \caption{Transition of the number of RBFs (averaged) generated within the CS-PNN for the CIL tasks using the seven datasets: a) isolet, b) letter-recognition, c) MNIST, d) optdigits, e) pendigits, f) sat, and g) segmentation.}
  \label{fig:sim22}
\end{figure}

Figures \ref{fig:sim21} and \ref{fig:sim22} show the transition of the CS-PNN and DNN classification accuracies and the number of RBFs generated within the CS-PNN for the CIL tasks using the seven datasets, respectively.  In Figs. \ref{fig:sim21} and \ref{fig:sim22}, the number $i$ ($i=1,2,3,4$) in the label 'task $i$' corresponds to the number of classes newly added; for task$1$, the initial number of classes was exceptionally $2$ to perform a two-class classification task, and each of the remaining partial data for a single class was continuously presented to the pattern classifiers.  Note that the classification accuracies in these two figures are those averaged over ten randomly generated permutations of the order of class data presentation to the pattern classifiers.  Also, the number of classes for the final task $N_{ft}$ was given as follows:

\begin{equation}
  N_{ft} = \mbox{rem}(N_c, i)~,
  \label{eqn:n_ft}
\end{equation}

where the function rem($x, y$) returns the remainder of the division $x/y$.
\begin{table}
 \caption{Summary of the simulation results obtained at each final stage of the CIL tasks}
  \centering
   \begin{tabular}{llccc}
    \toprule
     & & \multicolumn{2}{c}{CS-PNN} & DNN with iCaRL\\
    \cmidrule(r){3-4} \cmidrule(r){5-5}
    Dataset & Tasks & Av. $N_h$ & Av. Acc. (\%) & Av. Acc. (\%)\\
    \midrule
    isolet & task1 & 624 & 78.4 & \textbf{81.0}\\
    & task2 & 664 & \textbf{78.8} & 44.2\\
    & task3 & 731 & \textbf{79.8} & 30.8\\
    & task4 & 759 & \textbf{80.1} & 25.3\\
    \midrule
    letter-recognition & task1 & 1428 & \textbf{88.0} & 33.7\\
    & task2 & 1452 & \textbf{88.3} & 14.3\\
    & task3 & 1517 & \textbf{89.0} & 10.2\\
    & task4 & 1561 & \textbf{89.2} & 8.9\\
    \midrule
    MNIST & task1 & 4297 & \textbf{89.1} & 12.5\\
    & task2 & 4469 & \textbf{89.3} & 14.0\\
    & task3 & 4231 & \textbf{93.0} & 11.7\\
    & task4 & 3490 & \textbf{93.5} & 11.4\\
    \midrule
    optdigits & task1 & 116 & \textbf{91.7} & 79.4\\
    & task2 & 125 & \textbf{91.8} & 49.7\\
    & task3 & 132 & \textbf{93.1} & 36.8\\
    & task4 & 137 & \textbf{93.3} & 36.3\\
    \midrule
    pendigits & task1 & 200 & \textbf{92.1} & 77.0\\
    & task2 & 210 & \textbf{92.5} & 52.5\\
    & task3 & 228 & \textbf{92.9} & 35.0\\
    & task4 & 226 & \textbf{93.4} & 35.8\\
    \midrule
    sat & task1 & 207 & \textbf{73.5} & 59.2\\
    & task2 & 264 & \textbf{74.6} & 51.0\\
    & task3 & 284 & \textbf{75.4} & 52.4\\
    & task4 & 287 & \textbf{77.0} & 37.8\\
    \midrule
    segmentation & task1 & 31 & \textbf{78.8} & 35.5\\
    & task2 & 32 & \textbf{78.7} & 26.9\\
    & task3 & 38 & \textbf{80.2} & 22.8\\
    & task4 & 41 & \textbf{82.3} & 36.8\\
    \bottomrule
  \end{tabular}
  \label{table:sim21}
\end{table}

Although the iCaRL's classification rates were higher than the CS-PNN at each task's first partial data presentations, they dropped quickly during the simulations.  This quick drop is outstanding, especially where multiple classes were added as the CIL tasks proceeded (i.e., for tasks 2-4), as shown in Fig. \ref{fig:sim21} and Table \ref{table:sim21}.  On the other hand, the performance degradation remained relatively small overall for the CS-PNN compared to the iCaRL for all the cases.  For the CS-PNN, the number of RBFs accommodated constantly increased during the CIL tasks for each case (except for the MNIST), as shown in Fig. \ref{fig:sim22}, but remained smaller than that of the standard classification tasks at all the final stages (cf. Tables \ref{table:sim1} and \ref{table:sim21}).  Also, the comparison between Tables \ref{table:sim1} and \ref{table:sim21} reveals that the performance for the final stages, i.e., where all the classes were learned, was uniformly inferior to that for the standard tasks.  These observations imply the inherent difficulty of performing the CIL tasks compared with the standard ones.

Nevertheless, Table \ref{table:sim21} shows that the CS-PNN's performance gradually improved as the number of classes available for each task increased.  It indicates that the more classes become available, the more the CS-PNN can better estimate the pattern space.  In contrast to the CS-PNN, the performance obtained using the iCaRL degraded with the increasing number of classes per task; the performance degradation was getting severe at later tasks, as shown in Fig. \ref{fig:sim21}. This indicates that the DNN trained using the iCaRL failed to track the varying pattern space effectively and still was not able to avoid the catastrophic forgetting.
\subsection{Continuous Multi-Class Unlearning and Incremental Learning Tasks}
In the simulation study, another scenario was considered, where unlearning and incremental learning tasks are subsequently and repetitively imposed on the CS-PNN, namely continuous unlearning and incremental learning (CUIL).  Although a variety of CUIL tasks can be devised, we restricted to the following simulation setup in this work and investigated the impact on the performance under a dynamically varying condition:

\setcounter{bean}{1}
\begin{list}{\arabic{bean})}{\usecounter{bean}}
\item Initial setup: construct a CS-PNN using an entire training dataset (i.e., containing all the training data for all the classes) by applying \textbf{Algorithm \ref{alg-construction}}.  Test it (\textbf{Algorithm \ref{alg-testing}}).;
\item For $i=1$ to $N_{ds}$ do:
  \setcounter{subbean}{1}
  \begin{quote}
    \begin{list}{2-\arabic{subbean})}{\usecounter{subbean}}
    \item Unlearn the $N_{ul}$ classes arbitrarily 
      chosen from the CS-PNN (\textbf{Algorithm 
      \ref{alg-unlearning}}) and test it using an 
      unknown dataset for all the remaining classes 
      (\textbf{Algorithm \ref{alg-testing}});
    \item Perform a CIL task over the $N_{ul}$ classes 
      unlearned in the previous sub-step (i.e., 
      performing the reconstruction by applying 
      \textbf{Algorithm \ref{alg-construction}}) and 
      test the reconstructed CS-PNN using an unknown 
      dataset for all the remaining plus the $N_{ul}$ 
      classes (\textbf{Algorithm \ref{alg-testing}});
    \end{list}
  \end{quote}
\end{list}

In the simulation study, $N_{ds}=4$, and the number of classes $N_{ul}$ for each unlearning stage/CIL was  varied according to the setup:

\begin{equation}
  N_{ul} = \mbox{floor}(N_c/j)~,
  \label{eqn:n_ul}
\end{equation}

where $j = 2, 3,$ and $4$ and the function floor($x$) returns the largest integer given a scalar $x$.  The setting by (\ref{eqn:n_ul}) then determines the amount of structural change imposed on the CS-PNN during the CUIL tasks; i.e., the setting with $j = 2$ can cause a dramatic change in the network structure, and (about) half of the classes will be unloaded from the network and added again later.

Note that, with the simulation setup above, the resultant network configuration (i.e., in terms of both the centroid vectors and the number of hidden layer units) obtained at each iteration will differ from the previous one, even though the number of classes accommodated within the CS-PNN returns to the one in the initial setup by reusing the original training dataset.

\begin{figure}[h]
  \centering
  \includegraphics[width=\textwidth]{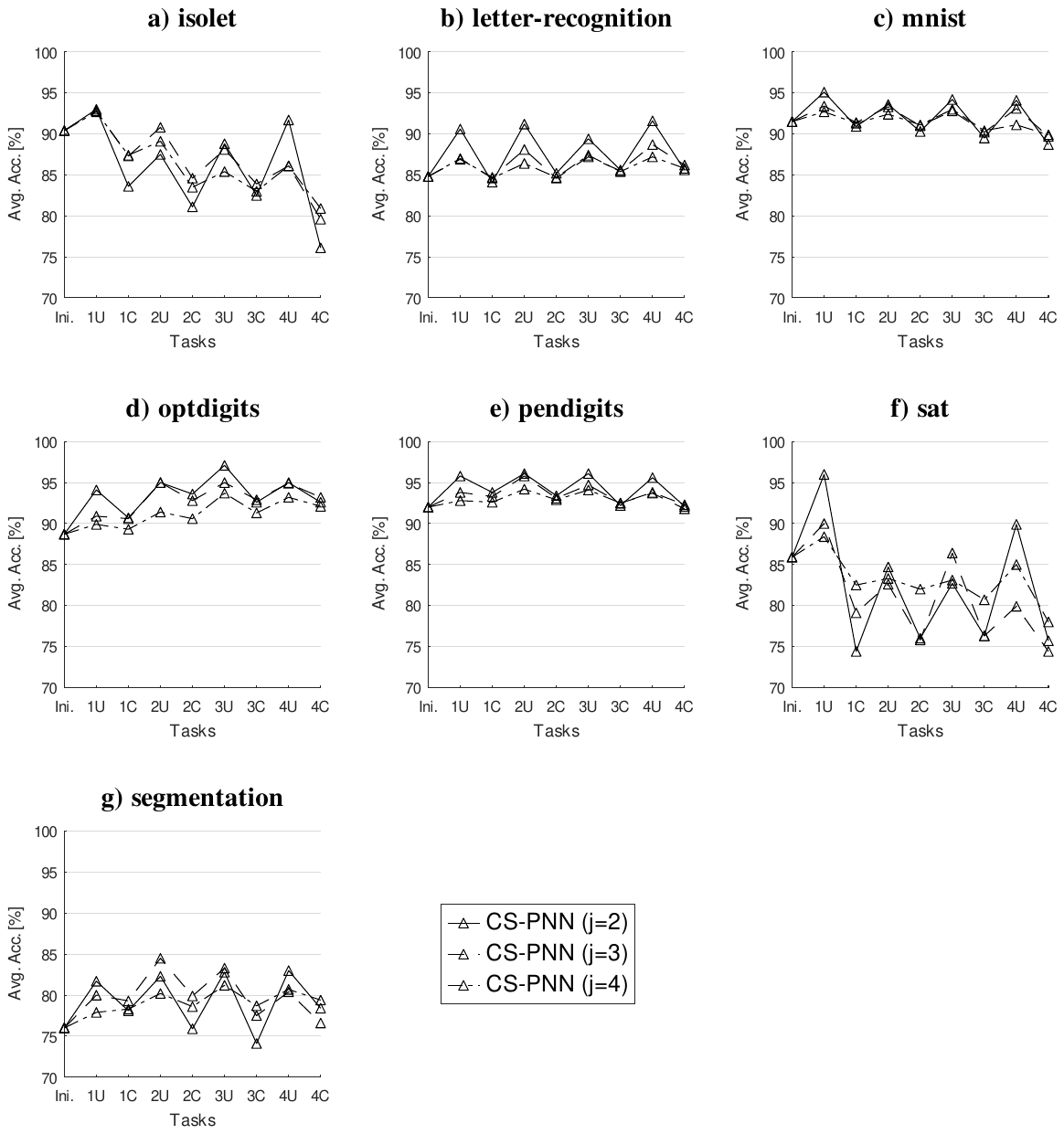}
  \caption{Transition of the classification accuracy (averaged) for the CUIL tasks using the seven datasets: a) isolet, b) letter-recognition, c) MNIST, d) optdigits, e) pendigits, f) sat, and g) segmentation.}
  \label{fig:sim31}
\end{figure}

\begin{figure}[h]
  \centering
  \includegraphics[width=\textwidth]{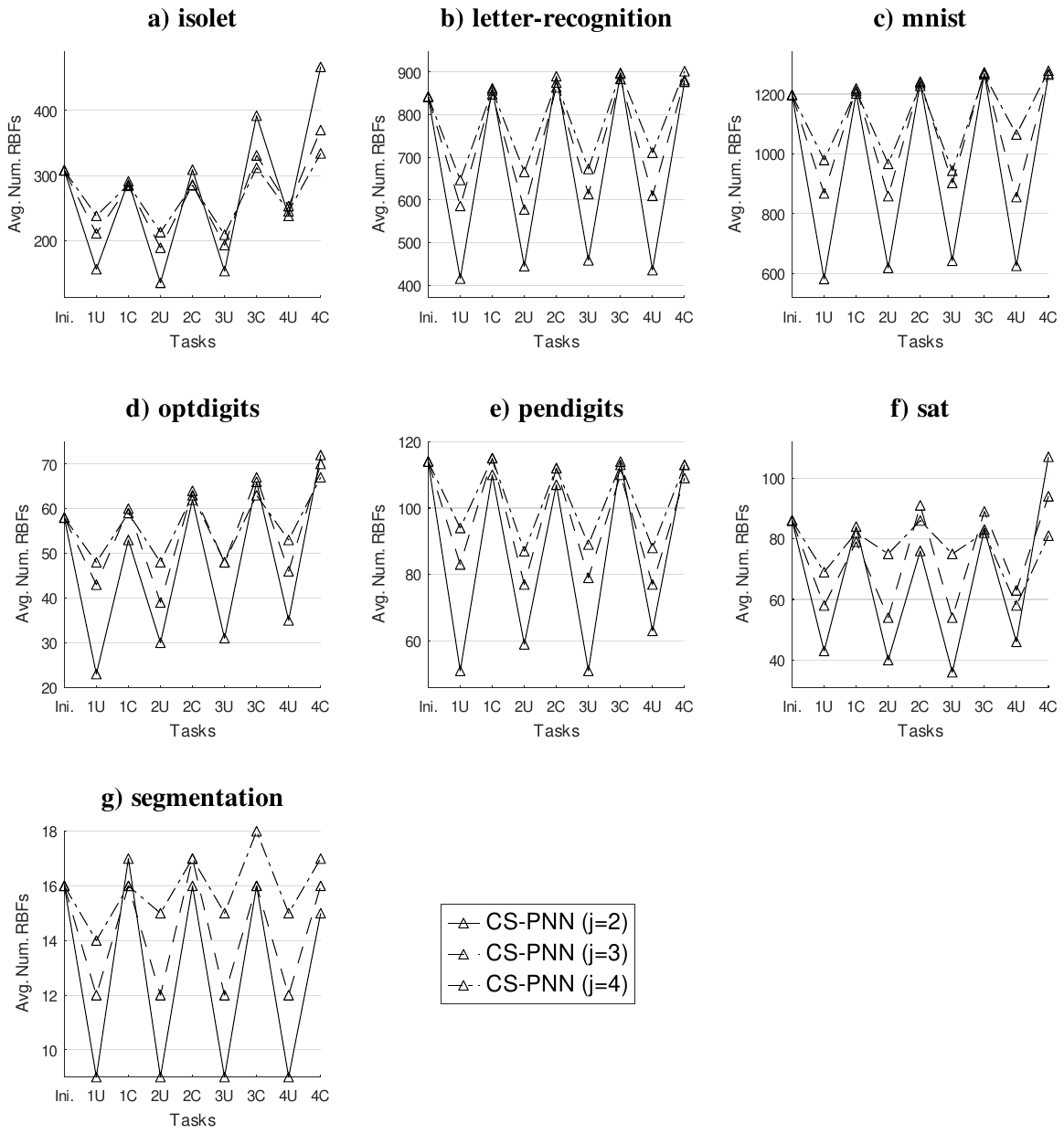}
  \caption{Transition of the number of RBFs (averaged) generated within the CS-PNN for the CUIL tasks using the seven datasets: a) isolet, b) letter-recognition, c) MNIST, d) optdigits, e) pendigits, f) sat, and g) segmentation.}
  \label{fig:sim32}
\end{figure}

Figures \ref{fig:sim31} and \ref{fig:sim32} show the simulation results of the CUIL tasks using the same seven datasets as used for the CIL tasks described in the previous subsection.

In these figures, the label on the $x$-axis 'Ini.' corresponds to the result obtained using the CS-PNN constructed with the whole training dataset (i.e., corresponding to the `Initial setup' in the above), whereas the labels, such as '2U' and '2C', denote those obtained after performing unlearning and a CIL at iteration $i=2$, respectively.  Similar to Figs. \ref{fig:sim21} and \ref{fig:sim22}, the results shown in Figs. \ref{fig:sim31} and \ref{fig:sim32} are those averaged over ten different runs, each randomly shuffling the class data, respectively.

As shown in Fig. \ref{fig:sim31}, the classification accuracy was constantly improved after each unlearning stage for all the cases, and the fluctuation in the classification accuracies between each unlearning and CIL stage became smaller with increasing the value $j$.  These observations may not correspond to the inherent nature of the CS-PNN acting as a pattern classifier but rather reflect that separating the entire pattern space was more straightforward with fewer classes.

On the other hand, the difference in the number of RBFs within the CS-PNN became more significant between these stages as the number of classes accommodated/unlearned increased, as in Fig. \ref{fig:sim32}; the number of RBFs was increased according to the degree of complexity in the pattern space to cover.  It is interesting to note, as shown in Fig. \ref{fig:sim32}, that the ratio of the number of RBFs after performing a CIL to that before it roughly coincides with that of $1$ to $1-1/j$.

While a large discrepancy was observed in the number of RBFs between class unlearning and CIL stages, as in Fig. \ref{fig:sim32}, the classification accuracy overall varied relatively little between the two stages, as in Fig. \ref{fig:sim31}.  In other words, the CS-PNN was able to cope reasonably well with the CUIL situations.  However, the classification accuracies at the last stage (i.e., at iteration $i=4$) for the a) isolet and f) sat cases considerably dropped from the beginning of the task, while the other five remained almost the same.  Therefore, it is considered that the pattern space separation for these two cases was more challenging than the other five.
\section{Conclusion}
In this paper, a novel approach to pattern classification based on compact-sized PNN capable of both class incremental learning and unlearning has been proposed.  Unlike many existing approaches, the training of the proposed CS-PNN does not require any hyperparameter tuning in advance or iterative network parameter approximation but is carried out by applying a simple, data-driven construction/reconstruction algorithm.  The network is dynamically reconfigured during the unlearning and incremental training stages; both the hidden and output layer units flexibly vary depending upon the situation.  This flexible network structure reconfiguration benefits from the inherent locality nature of the PNN model.

In the simulation study, the performance of the proposed CS-PNN approach has been evaluated under three different task scenarios, i.e., i) standard classification, ii) class incremental learning, and iii) continuous multi-class unlearning and incremental learning tasks.

For scenario i), it has been shown that the CS-PNN can yield a reasonable classification performance, while the number of RBFs accommodated within the network is much less than that of the original PNN approach.  In contrast, the overall performance was more or less comparable to that obtained using DNNs, though the DNNs could attain it with a smaller number of hidden layer units.  Despite this, the analytical study \cite{Takahashi-2022} suggests that implementing it in a parallel computing environment can alleviate the drawbacks inherent to the PNN models in the reference (i.e., testing) mode, provided that a sufficient memory resource is available for the RBFs; the computation time in the testing mode in the parallel setup can be as fast as that of a DNN with a fewer number of the hidden units, which is, therefore, currently under investigation for the CS-PNN case.

On the other hand, the DNNs did not work sufficiently well for ii); the DNN trained using the iCaRL significantly degraded with an increasing number of new classes in the CIL situations, probably showing the catastrophic forgetting phenomenon.  In contrast, the CS-PNN using the proposed reconstruction scheme did not suffer from such a problem, and a reasonable classification performance at each task was maintained.  Moreover, scenario iii) has shown that the CS-PNN can perform multiple subsequent class unlearning and incremental learning tasks without manipulating hyperparameters nor fiddling with other algorithm- and/or model-specific issues, unlike many other existing artificial neural network models.  Further, the empirical evidence given in this work supports the promising vision that the CS-PNN can provide a rapid and flexible pattern recognition engine for developing higher-level intelligent processing systems.

Future work is directed to investigate cases with more classes as well as larger databases than those used in this study.

%
\end{document}